\documentclass[sigconf, screen]{acmart}

\AtBeginDocument{%
  }

\copyrightyear{2026}
\acmYear{2026}
\setcopyright{cc}
\setcctype{by}
\acmConference[DAC '26]{63rd ACM/IEEE Design Automation Conference}{July 26--29, 2026}{Long Beach, CA, USA}
\acmBooktitle{63rd ACM/IEEE Design Automation Conference (DAC '26), July 26--29, 2026, Long Beach, CA, USA}
\acmDOI{10.1145/3770743.3804115}
\acmISBN{979-8-4007-2254-7/2026/07}

\usepackage{hhline}
\usepackage{amsmath,mathtools,bm}
\usepackage{graphicx}
\usepackage{booktabs}
\usepackage{tabularx}
\usepackage[flushleft]{threeparttable}
\usepackage{mathrsfs}
\usepackage{xcolor}
\usepackage[linesnumbered,ruled,vlined]{algorithm2e}
\usepackage{multirow}
\usepackage{enumitem}
\usepackage{pifont}

\newcommand{\tworows}[1]{\multirow{2.3}{*}{#1}}

\newcommand{\eq}{\!=\!}

\newcommand{\CC}{\mathbf{C}}

\begin{document}

\title{AttentionCap: Transformer Based Capacitance Matrix Learning Toward Full-Chip Extraction}

\author{Jiechen Huang$^1$, Hector R. Rodriguez$^1$, Dingcheng Yang$^1$, Zuochang Ye$^2$, Yibo Lin$^3$, Wenjian Yu$^{1*}$}
\thanks{This work is supported by National Science and Technology Major Project (2021ZD0114703) and the MIND project (MINDXZ202406). $^*$Corresponding author.}
\affiliation{
$^1$Dept. Computer Science \& Tech., BNRist, Tsinghua Univ., Beijing, China;\\ $^2$School of IC, BNRist, Tsinghua Univ., Beijing, China; $^3$School of IC, Peking Univ., Beijing, China.
\country{}
}

\renewcommand{\shortauthors}{Huang et al.}

\begin{abstract}

As capacitance extraction accuracy of rule-based pattern matching becomes difficult to sustain at advanced nodes, 
a growing trend emerges to develop deep-learning-based 2D capacitance models. 
However, existing MLP- and CNN-based methods constrain their input to fixed metal-layer combinations in a specific process node, limiting their usability in practice.
Recognizing the inherent similarity between capacitance matrix and the prevailing attention mechanism, we propose AttentionCap, a customized Transformer for capacitance matrix learning, with a Gram representation framework, a physics-aligned symmetric-attention output layer, and a novel normalized Laplacian loss. We also introduce a process-node embedding to enable multi-node learning. 
Trained on synthetic data, AttentionCap attains 0.67\%/3.99\% self/coupling-capacitance error on unseen real designs under a multi-layer and multi-node setting, surpassing the CNN-Cap baseline with 4.6$\times$/5.7$\times$ lower self/coupling error and 192$\times$ faster inference speed.
A pretrained AttentionCap accurately transfers to an unseen node with only 5K samples and 4K finetuning steps. 
With sufficient accuracy on unseen real designs and strong transferability to new process nodes, AttentionCap offers highly practical value for modern EDA workflows. Code and data are available at \url{https://github.com/THU-numbda/AttentionCap}.

\end{abstract}

\keywords{Capacitance Extraction, Cross-Sectional Pattern Matching, Transformer, Attention Mechanism}

\maketitle

\section{Introduction}

Parasitic capacitance extraction solves the electric coupling among interconnects in integrated circuits (ICs) and delivers a realistic circuit model for accurate timing/signal-integrity/power analysis \cite{chen2014interconnect}. 
It has become a pronounced bottleneck at advanced process nodes, as the accuracy of pattern matching methods becomes difficult to maintain, while accurate field solvers face scalability challenges with the ever-growing design size and complexity \cite{yu2021advancements}. 

Full-chip capacitance extraction uses pattern matching method with 2D cross-sectional pattern libraries, which match 2D slices of layouts to predefined patterns or interpolation formulas \cite{cong1997}.
These libraries require tedious manual design and tuning to maintain accuracy for different process nodes, and are often proprietary,
which has motivated the use of deep learning as a compelling alternative \cite{yu2025deep}. 
CNN-Cap \cite{yang2021cnn} has established the paradigm of using density grids to represent 2D conductor geometries \cite{abouelyazid2022accuracy,tsai2025rescap,yang2023cnn}. It provides distance-aware input for convolutional neural networks (CNNs) to learn the capacitance among conductors. 
Some studies also adopted coordinate-based input for multilayer perceptrons (MLPs) \cite{abouelyazid2022fast, ma2023extraction, yu2025ail}.
However, existing 2D capacitance models either admit only a fixed number of conductors \cite{ma2023extraction,yu2025ail} or output only the capacitance of one conductor pair \cite{abouelyazid2022fast, abouelyazid2022accuracy, yang2021cnn,tsai2025rescap}, limited by the fixed-dimension nature of MLPs or CNNs. 
More critically, they exhibit limited generalizability because the model is trained for each metal-layer combination (typically 3 layers like M1-M3-M5) \cite{yang2021cnn,abouelyazid2022accuracy,abouelyazid2022fast, ma2023extraction, yu2025ail,tsai2025rescap},
thus requiring numerous datasets and retrained models to cover a process node.

\begin{table}[t]
  \setlength{\tabcolsep}{3pt}
\setlength{\abovecaptionskip}{2pt}
\setlength{\belowcaptionskip}{2pt}
  \caption{AttentionCap vs. CNN-Cap-like methods.}
  \label{tab:comp}
  \begin{threeparttable}
  \begin{tabularx}{\columnwidth}{@{}lll@{}}
    \toprule
    & CNN-Cap-like \cite{yang2021cnn,abouelyazid2022accuracy,tsai2025rescap}
    & AttentionCap \\
    \midrule
    Input represent. & Density grids & Coordinates \\
    Input conductors & $n$ (variable) & $n$ (variable) \\
    Output dimension & Scalar (total or coupling) & \textbf{$n{\times}n$ (full matrix)} \\
    Full extraction  & $n^2$ inference passes & \textbf{1 inference pass} \\
    Layer combination & Fixed (3 layers) & \textbf{Any} \\
    Process coverage\tnote{$\ast$} & $\binom{m}{3}$ datasets/models & \textbf{1 dataset/model} \\
    Multi-Node & Not explored & \textbf{1 model} \\
    \bottomrule
  \end{tabularx}
  \begin{tablenotes}[flushleft]
    \small
    \item[$\ast$] ``Process coverage'' refers to the necessary datasets and retrained models to cover all 3-layer combinations in an $m$-layer process node.
  \end{tablenotes}
  \end{threeparttable}
\end{table}

The Transformer architecture \cite{vaswani2017attention} has achieved state-of-the-art results with remarkable generalizability across domains such as large language models (LLMs) \cite{devlin2019bert}, computer vision \cite{dosovitskiy2021an}, and IC design \cite{wen2022layoutransformer, kim2024traceformer}.
At its core, the self-attention mechanism learns the pairwise relations among $n$ inputs via an $n\times n$ \textit{attention matrix}, 
and accordingly refines the representation of each input.
By effectively capturing complex relations, attention mechanism endows Transformers with strong representational power, making it the \textit{de facto} foundation of modern AI \cite{achiam2023gpt}.

Recognizing the structural similarity that both capacitance matrix and attention mechanism capture an $n$-to-$n$ interaction,
we propose \textit{AttentionCap}, which leverages the prevailing Transformer architecture for capacitance matrix learning toward full-chip extraction. 
While prior studies mainly focus on individual capacitance elements \cite{abouelyazid2022fast, abouelyazid2022accuracy, yang2021cnn,tsai2025rescap}, 
the $n\times n$ \textit{capacitance matrix} fully characterizes an $n$-conductor layout,
quantifying the coupling strength between each conductor pair. 
This relation is determined not only by their distance but also by other conductors 
and dielectric environment, 
which inherently limits distance-based representations such as density grids in CNN-Cap-like methods \cite{yang2021cnn} and 3D-distance-based graphs in GNN-Cap for 3D extraction \cite{liu2023gnn}.
In AttentionCap, we input a conductor sequence, train a Transformer to map each conductor to a ``capacitance embedding'', and dot-product them via symmetric attention to produce the full capacitance matrix.
As summarized in Table~\ref{tab:comp}, AttentionCap inherits several key advantages from Transformer: 
natively support for variable-length inputs, uniform all-to-all interactions without receptive-field limits, and one-pass construction of the full matrix.
It is also surprisingly generalizable: one model can accurately cover all metal layers and even multiple process nodes.
Our main contributions and results are as follows:

\begin{itemize}[leftmargin=*]
    \item Based on matrix decomposition and Gram representation framework, we propose AttentionCap, a Transformer specialized toward capacitance physics: order equivariance, a symmetric attention layer enforcing Laplacian structure, and a normalized Laplacian loss for stable training.

    \item We introduce learnable process-node embeddings for multi-node learning, significantly enhancing AttentionCap's accuracy and new-node transferability, which is highly valuable in practical EDA workflows.

    \item Through extensive experiments on synthetic training data and unseen real-design test sets, we show that AttentionCap achieves 1.25\% and 6.26\% errors for self- and coupling-capacitance on unseen designs (averaged on 7nm and 65nm) under the challenging multi-layer setting, largely outperforming CNN-Cap in accuracy and efficiency. Under multi-node training, these errors are further reduced to 0.67\% and 3.99\%, and the pretrained AttentionCap shows strong transferability to new nodes with only 5K samples and 4K finetuning steps.
\end{itemize}

\section{Preliminaries}
\subsection{Interconnect Capacitance Extraction}
\label{sec:prelim_cap}

For a system of $n$ conductors, the relation between their electric potentials $\mathbf{u} \in \mathbb{R}^{n}$ and electric charges $\mathbf{q} \in \mathbb{R}^{n}$ is described by the \textit{capacitance matrix} $\CC \in \mathbb{R}^{n \times n}$ via a linear equation \cite{maxwell1873treatise}
\begin{equation}
    \mathbf{q} = \CC \mathbf{u}.
\end{equation}
The diagonal element $C_{ii}$ is the \textit{total capacitance} of conductor $i$, and $C_{ij}$ for $i \ne j$ is the \textit{coupling capacitance} between $i$ and $j$. $\CC$ satisfies important physical properties \cite{huang2025parallel}: (1) sign property, $C_{ii} \!\ge\! 0$ and $C_{ij} \!\le\! 0$ for $i \ne j$; (2) symmetry, $C_{ij}\!=\!C_{ji}$; (3) zero row-sum, $\sum_{j=1}^{n} C_{ij}\!=\! 0$. 
For ease of presentation, we use an equivalent form where couplings are defined as $|C_{ij}|$, so that closer conductors intuitively correspond to larger couplings.
Hence, $\CC$ is a signless \textit{Laplacian matrix} \cite{merris1994laplacian} and is guaranteed to be positive semi-definite. 

 \textit{Field solvers} or \textit{pattern matching} is employed in interconnect capacitance extraction to solve $\CC$. 
Field solvers simulate the electrostatic field with finite difference method (FDM), finite element method (FEM), boundary element method (BEM) \cite{yu2009variational} or floating random walk (FRW) method \cite{huang2024enhancing}. 
While being accurate, they are computationally expensive and cannot scale to full-chip extraction. Pattern matching is an approximate approach for full-chip extraction, which decomposes the 3D layout into 2D cross-sections and estimates capacitance with look-up tables \cite{cong1997}. 

\subsection{Deep-Learning-Based Capacitance Extraction for 2D Cross-Sections}
Since the design and usage of pattern libraries are highly empirical and not transparent to the research community, there is a growing trend of deep-learning-based 2D cross-sectional capacitance modeling \cite{yu2025deep}.
They can be divided into two categories:
(1) \textbf{Conductor coordinates as input} were adopted by MLP-based models 
for process-variation-aware extraction \cite{abouelyazid2022fast}, capacitance matrix extraction \cite{ma2023extraction}, and adaptive incremental learning \cite{yu2025ail}. However, they either supported only a fixed number of conductors \cite{li2020layout,ma2023extraction, yu2025ail} or predicted only one pairwise coupling \cite{abouelyazid2022fast}.
(2) \textbf{Density-grid encoding as input} \cite{abouelyazid2022accuracy,yang2021cnn} enabled flexibility for variable conductor counts. 
CNN-Cap \cite{yang2021cnn} leveraged ResNet to process density grids with high capacitance accuracy.
ResCap \cite{tsai2025rescap} combined conductor-length-based estimation and density-grid MLPs for standard cell designs.
However, density-grid models must mark the targets in the grid, and predict only the targeted capacitance. 
For an $n$-conductor layout, the grid instantiations and model inference must repeat $n^2$ times.
Moreover, existing models share a critical limitation: they are trained on fixed combinations of (typically) 3 metal layers, necessitating at least $\binom{m}{3}$ separate datasets and retrained models to cover an $m$-layer process node.

\subsection{Attention Mechanism and Modern Transformer Architecture}\label{sec:prelim_att}
\begin{figure}[t]
    \centering
\setlength{\abovecaptionskip}{2pt}
\setlength{\belowcaptionskip}{2pt}
    \includegraphics[width=0.8\linewidth]{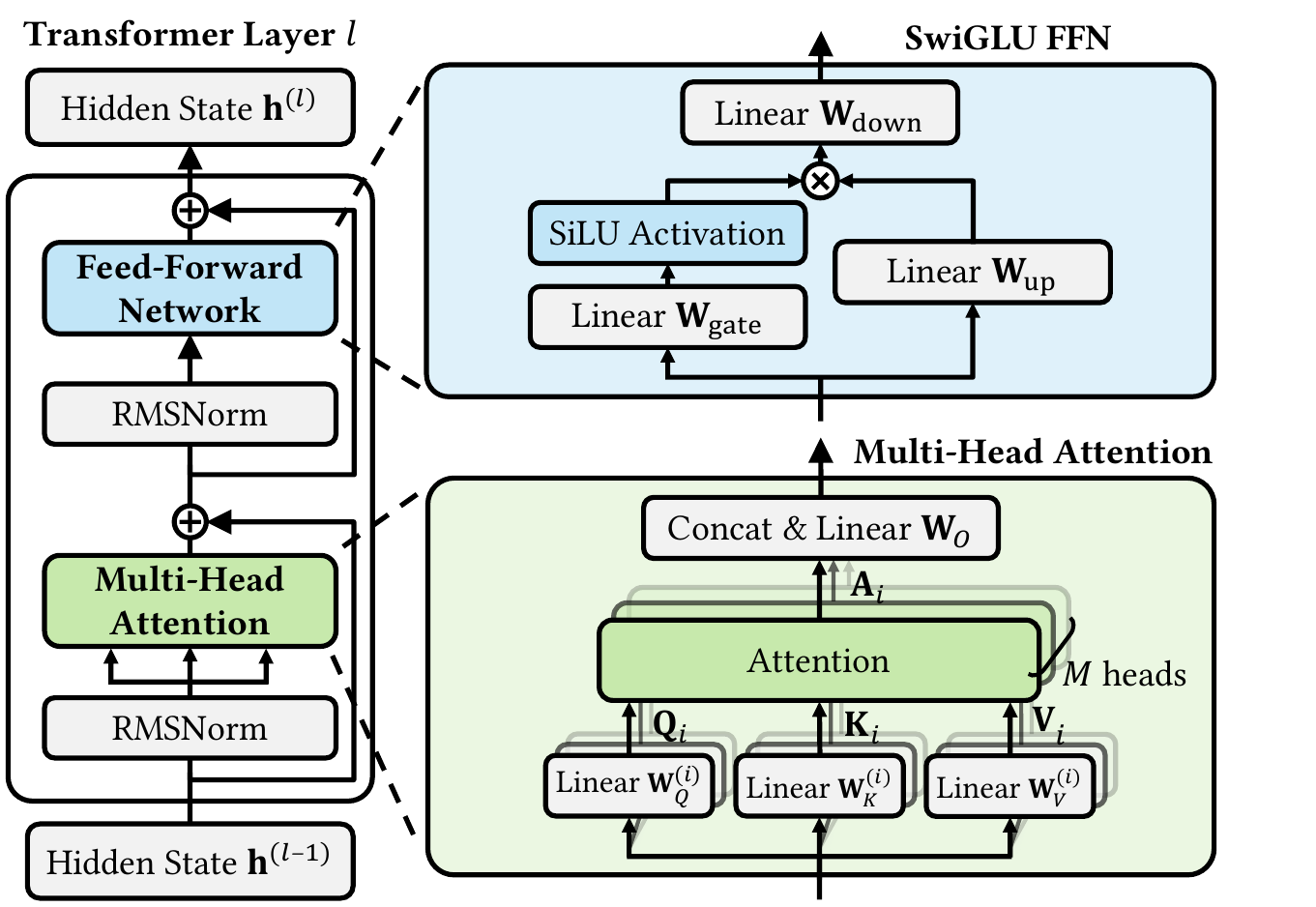}
    \caption{Modern Transformer architecture.}
    \label{fig:arch}
\end{figure}
\begin{figure*}[t]
    \centering
\setlength{\abovecaptionskip}{0pt}
\setlength{\belowcaptionskip}{0pt}
    \includegraphics[width=\textwidth]{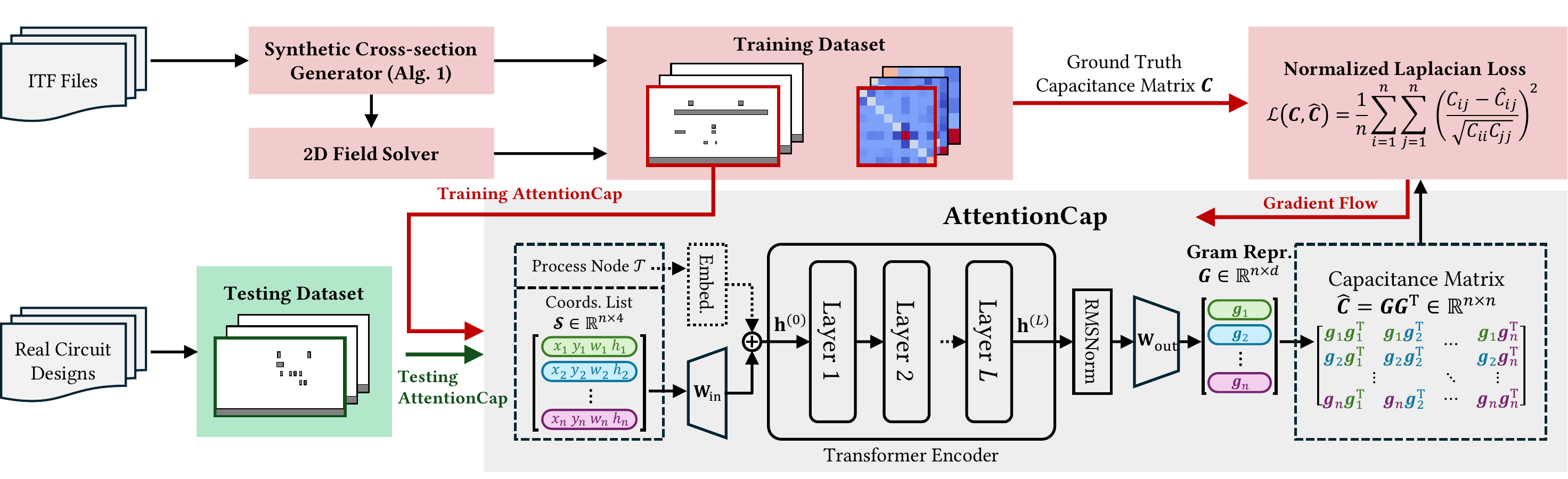}
    \caption{Overview of the AttentionCap framework for capacitance matrix learning.}
    \label{fig:overview}
\end{figure*}
We briefly review the Transformer architecture \cite{vaswani2017attention}, along with LLM-era enhancements used in this work, e.g., SwiGLU \cite{shazeer2020glu} and RMSNorm \cite{zhang2019root}.
For an input sequence of length $n$, an $L$-layer Transformer maps the input embedding $\mathbf{h}^{(0)}\!\in\!\mathbb{R}^{n\times d}$ to a final representation $\mathbf{h}^{(L)}\!\in\!\mathbb{R}^{n\times d}$, where $d$ is the model dimension. As illustrated in Fig. \ref{fig:arch}, each layer transforms $\mathbf{h}^{(l-1)}\!\mapsto\!\mathbf{h}^{(l)}$ via a multi-head self-attention (MHA) module followed by a position-wise feed-forward network (FFN). Each attention head $i$ uses weight matrices $\mathbf{W}_Q^{(i)},\mathbf{W}_K^{(i)},\mathbf{W}_V^{(i)}\!\in\!\mathbb{R}^{d\times d_k}$ and MHA concatenates and projects the outputs across $M$ heads with weight $\mathbf{W}_O\!\in\!\mathbb{R}^{Md_k\times d}$:
\begin{equation}\label{eq:att}
    \mathbf{O}_i \eq \mathsf{softmax}\!\left(\frac{\mathbf{Q}_i\mathbf{K}^{\top}_i}{\sqrt{d_k}}\right)\mathbf{V}_i,\ \ \     \mathsf{MHA}(\mathbf{h})\eq [\mathbf{O}_1,\ldots,\mathbf{O}_M]\mathbf{W}_O,
\end{equation}
where $\{\mathbf{Q}_i,\mathbf{K}_i,\mathbf{V}_i\}\eq \{\mathbf{h}\mathbf{W}_{Q}^{(i)}, \mathbf{h}\mathbf{W}_K^{(i)},\mathbf{h}\mathbf{W}_V^{(i)}\}$.
In modern architectures, the FFN commonly adopts the SwiGLU form \cite{shazeer2020glu}:
\begin{equation}
    \mathsf{FFN}(\mathbf{x}) = \Big[(\mathbf{x}\mathbf{W}_{\text{up}} ) \otimes \sigma(\mathbf{x}\mathbf{W}_{\text{gate}})\Big] \mathbf{W}_{\text{down}}^{\top},
\end{equation}
where $\mathbf{W}_{\text{up}},\mathbf{W}_{\text{gate}},\mathbf{W}_{\text{down}}\!\in\!\mathbb{R}^{d\times d_{\text{ff}}}$ are weights, $\otimes$ is element-wise product, and $\sigma$ is the SiLU activation \cite{shazeer2020glu}.
Each MHA or FFN module is preceded by RMSNorm \cite{zhang2019root}:
\begin{equation}
    \mathsf{RMSNorm}(\mathbf{x}) = \frac{\mathbf{x}}{\sqrt{\tfrac{1}{d}\!\sum_{i=1}^d\!x_i^2+\epsilon}} \otimes \bm{\gamma},
\end{equation}
where $\bm{\gamma}\!\in\!\mathbb{R}^d$ is learnable and $\epsilon$ is for numerical stability.

\section{AttentionCap: Capacitance Transformer}

\subsection{Problem Formulation}

We formulate \textit{capacitance matrix learning} as follows. A layout cross-section with $n$ conductors is parameterized as sequence $\bm{\mathcal{S}}= [(x_1, y_1, w_1, h_1), \dots, (x_n, y_n, w_n, h_n)] \in \mathbb{R}^{n\times 4}$, where $x_i,y_i,w_i,h_i$ are the center coordinates, width, and height of the $i$-th conductor\footnote{Following prior studies, we consider only rectangular Manhattan conductors. AttentionCap could potentially accommodate non-Manhattan geometries with different input embeddings; a systematic study is left for future work.}. 
For a process technology node $\mathcal{T}$, the dielectric stack is fixed (specified in an interconnect technology file (ITF)), so our target is a deterministic but highly non-linear mapping $\mathcal{F}_{\mathcal{T}}$:
\begin{equation}
    \mathcal{F}_{\mathcal{T}}: \mathbb{R}^{n\times 4} \rightarrow \mathbb{R}^{n\times n},\quad \CC = \mathcal{F}_{\mathcal{T}}(\bm{\mathcal{S}}).
\end{equation}
Beyond this, we also explore a more general setting where multi-process-node $\mathcal{F}_{\mathcal{T}_1}, \mathcal{F}_{\mathcal{T}_2},\dots $ are learned jointly with a single model.

\subsection{Gram Representation Learning}
A common strategy for relation modeling is to learn latent embeddings and express the relations via their dot-product \cite{cicchetti2025gramian}, which is also the basic idea of attention \cite{vaswani2017attention}.
We follow this paradigm to decompose learning objective via dot-product, and design AttentionCap under a representation learning framework.

A key observation is that due to the symmetric positive semi-definite nature of capacitance matrix (see Section \ref{sec:prelim_cap}), $\mathbf{C}$ can be factorized by dot-product \cite{horn2013matrix}, as stated in Lemma \ref{lemma:gram}.
\begin{lemma} (Gram Representation) \label{lemma:gram}
    For any positive semi-definite matrix $\CC\in \mathbb{R}^{n\times n}$, there exists $\mathbf{G} \in \mathbb{R}^{n\times n'}$ such that 
    \begin{equation}
        \CC = \mathbf{G}\mathbf{G}^{\top},\quad \text{i.e.},\ C_{ij} = \langle\bm{g}_i,\bm{g}_j\rangle, \forall i,j,
    \end{equation}
    where $n' \le n$ is the rank of $\CC$ \cite{horn2013matrix}. We refer to $\mathbf{G}$ and its row vectors $\{\bm{g}_i\in \mathbb{R}^{1\times n'}\}$ as the Gram representation of $\mathbf{C}$.
\end{lemma}
With this decomposition, the target mapping becomes $\mathcal{G}_{\mathcal{T}}$:
\begin{equation}
    \mathcal{G}_{\mathcal{T}}: \mathbb{R}^{n\times 4} \rightarrow \mathbb{R}^{n\times d},\quad  \mathcal{F}_{\mathcal{T}}(\bm{\mathcal{S}}) = \mathcal{G}_{\mathcal{T}}(\bm{\mathcal{S}}) \mathcal{G}_{\mathcal{T}}(\bm{\mathcal{S}})^{\top}.
\end{equation}
This is a sequence-to-sequence task that transforms each conductor into a Gram vector or ``capacitance embedding''. 
We implement $\mathcal{G}_{\mathcal{T}}$ with a Transformer and customized input and output layers:
\begin{equation}\label{eq:overall}
    \mathbf{h}^{(0)} \!=\! \mathsf{in}(\bm{\mathcal{S}}),\ \mathbf{h}^{(L)}\!=\! \mathsf{Transformer}(\mathbf{h}^{(0)}),\ \hat{\mathbf{G}} \!=\! \mathsf{out}(\mathbf{h}^{(L)}).
\end{equation}

\subsection{Model Architecture}
Our methodology is outlined in Fig.~\ref{fig:overview}.
We propose several physics-aligned specializations based on standard Transformer: no positional encoding, no causal masking, additional input and output layers, and a normalized Laplacian loss.

\textbf{Input Embedding}. The $\mathsf{in}(\cdot)$ in Eq. \eqref{eq:overall} lifts the conductor coordinates $\bm{\mathcal{S}}\in \mathbb{R}^{n\times 4}$ to initial embedding $\mathbf{h}^{(0)}\in \mathbb{R}^{n\times d}$. We implement it as a position-wise linear layer with weight matrix $\mathbf{W}_{\text{in}}\in \mathbb{R}^{4\times d}$:
\begin{equation}\label{eq:input}
    \mathbf{h}^{(0)} = \bm{\mathcal{S}}\mathbf{W}_{\text{in}}.
\end{equation}

\textbf{No Positional Encoding}. Unlike language tasks, capacitance learning is agnostic to the order of input sequence. More formally, the target mapping $\mathcal{G}_{\mathcal{T}}$ is inherently permutation-equivariant:
\begin{equation}
    \mathcal{G}_{\mathcal{T}}(\pi\bm{\mathcal{S}}) = \pi\mathcal{G}_{\mathcal{T}}(\bm{\mathcal{S}}),\quad \forall \text{ permutation } \pi.
\end{equation}
Without positional encoding, the Transformer architecture natively satisfies this equivariance \cite{vaswani2017attention}. Therefore, we do not add any positional encoding in AttentionCap.
 
\textbf{Transformer Encoder}. The $\mathsf{Transformer}(\cdot)$ in Eq. \eqref{eq:overall} is the $L$-layer Transformer with modern enhancements, as described in Section~\ref{sec:prelim_att}.
We adopt the encoder-only architecture, since causal masking is unnecessary in our task.

\textbf{Symmetric-Attention Output Layer}. The $\mathsf{out}(\cdot)$ in Eq. \eqref{eq:overall} is implemented as an RMSNorm layer followed by a linear layer with weight $\mathbf{W}_{\text{out}} \in\mathbb{R}^{d\times d}$. The capacitance matrix prediction is then computed by dot-product:
\begin{equation} \label{eq:out}
    \hat{\mathbf{G}} = \mathsf{RMSNorm}(\mathbf{h}^{(L)}) \mathbf{W}_{\text{out}},\ \     \hat{\CC} = \frac{1}{\sqrt{d}}\hat{\mathbf{G}}\hat{\mathbf{G}}^{\top}.
\end{equation}
The scaling factor $\frac{1}{\sqrt{d}}$ follows attention mechanism Eq. \eqref{eq:att} to stabilize training. 
Note that Eq. \eqref{eq:out} is equivalent to \textit{symmetric attention} \cite{courtois2024symmetric}, so the output layer can be viewed as a symmetric attention module which outputs the attention matrix as the capacitance prediction.
Finally, to enforce the Laplacian properties of $\hat{\CC}$ (see Section \ref{sec:prelim_cap}), we set $\hat{C}_{ii} = \sum_{j\ne i}\hat{C}_{ij}, \forall i$.

\textbf{Normalized Laplacian Loss}. 
Since capacitance values can span across multiple scales, we propose a novel loss function for capacitance matrix learning. We apply the symmetric Laplacian normalization based on ground truth $\CC$:
\begin{equation}
    \CC'=\mathbf{D}^{-\frac{1}{2}}\CC\mathbf{D}^{-\frac{1}{2}},\quad \hat{\CC}'=\mathbf{D}^{-\frac{1}{2}}\hat{\CC}\mathbf{D}^{-\frac{1}{2}},
\end{equation}
where $\mathbf{D}$ is the diagonal matrix with $D_{ii}=C_{ii}$. The normalized Laplacian loss is defined as:
\begin{equation}
    \mathcal{L}(\CC, \hat{\CC}) = \frac{1}{n}\|\CC'-\hat{\CC}'\|_F^2 = \frac{1}{n}\sum_{i=1}^n \sum_{j=1}^n \left(\frac{C_{ij} - \hat{C}_{ij}}{\sqrt{C_{ii}C_{jj}}}\right)^2,
\end{equation}
where $\|\cdot\|_F$ is the Frobenius norm. This loss prevents domination by large capacitance values, and is shown to be crucially effective in experiments (see Table~\ref{tab:ablation}).

\begin{figure}
\setlength{\abovecaptionskip}{0pt}
\setlength{\belowcaptionskip}{0pt}
    \centering
    \includegraphics[width=0.9\linewidth]{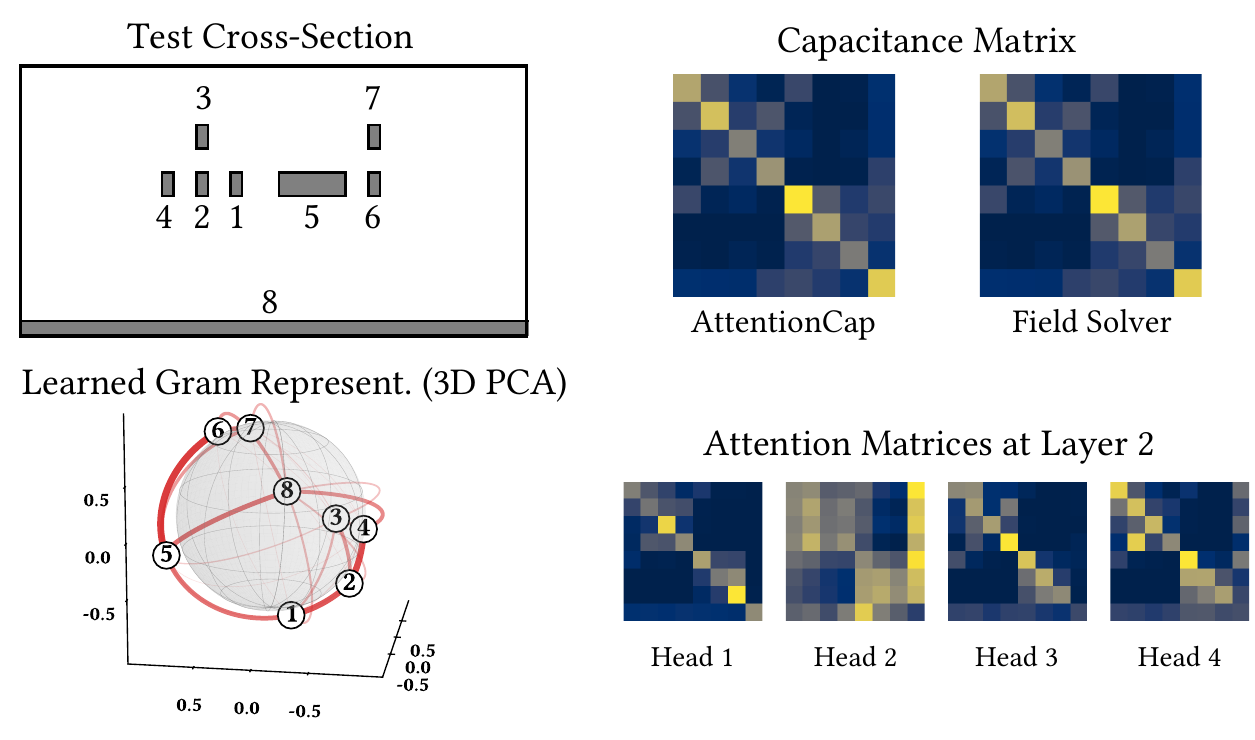}
    \caption{AttentionCap learns meaningful Gram representations for each conductor (visualized via 3D PCA on a unit sphere, line thickness reflecting their dot-products); its internal attention patterns match the target capacitance matrix.}
    \label{fig:attention}
\end{figure}

With the Gram representation framework and our architectural enhancements, AttentionCap effectively learns meaningful ``capacitance embeddings'' with attention patterns that align closely with the target capacitance structure, as shown in Fig.~\ref{fig:attention}.

\subsection{Multi-Process-Node Learning}
Building capacitance models for new process nodes is often a bottleneck in IC design practice, so a reusable multi-node model could be highly valuable.
AttentionCap not only supports multi-node learning, but also benefits from it in two ways: (1) more abundant and diverse samples in the mixed dataset, and (2) strong transferability due to separation of process-node-agnostic features.

\textbf{Process-Node Embedding}.
For process nodes $\mathcal{T}_1, \mathcal{T}_2,\dots$, their capacitance mappings $\mathcal{F}_{\mathcal{T}_1}, \mathcal{F}_{\mathcal{T}_2},\dots$ are governed by the different dielectric stacks and wire features.
We propose to use a learnable embedding to capture this node-specific information, and add it to the coordinate embedding in Eq. \eqref{eq:input}: 
\begin{equation}
    \mathbf{h}^{(0)} = \bm{\mathcal{S}}\mathbf{W}_{\text{in}} + \mathsf{Embed}(\mathcal{T}_i).
\end{equation}
This enables AttentionCap to differentiate samples from each node and implicitly inject dielectric-related information into the conductor representations. This extension is lightweight yet effective, as validated in experiments (see Table~\ref{tab:main}).

\textbf{Transfer to New Process Nodes}.
Trained on multi-node data, AttentionCap learns both 
process-node embeddings and process-node-agnostic features, developing strong transferability to new, unseen nodes.
For each new process node $\mathcal{T}_{\text{new}}$, we can add an embedding vector $\mathsf{Embed}(\mathcal{T}_{\text{new}})$ and finetune the pretrained model. Experiments show that a multi-node-pretrained AttentionCap can transfer to $\mathcal{T}_{\text{new}}$ with a quick finetune on few training samples from $\mathcal{T}_{\text{new}}$ (see Fig.~\ref{fig:transfer}). 
This fast adaption is exceptionally valuable in practice, as preparing training data with field solver is the main bottleneck for developing capacitance models for a new process node.

\subsection{Synthetic Training Data Generation}
Layout data is often scarce and proprietary. 
More critically, real layouts contain deterministic features such as wire width and layer usage, which may cause overfitting and poor generalizability without careful data curation.
Therefore, we propose a synthetic cross-section generator (Alg.~\ref{algo:synthetic}) that produces diverse training samples, requiring only a process node description (typically an ITF).

Alg. \ref{algo:synthetic} first randomly selects 3, 4, or 5 layers (line~1) and then places a random number of conductors. 
We use Poisson distribution $P(\cdot)$ to sample the conductor count $n$ (line~2). $\mathcal{N}(\cdot,\cdot)$ and $U\{\cdot\}/(\cdot)$ denote Gaussian and discrete/continuous uniform distributions, respectively.
To mimic realistic cross-sections, each conductor is sampled as ``continuous'' with 10$\%$ probability (line~9) or ``discrete'' with 90$\%$ probability (lines~11-12).
For discrete width sampling, we apply the exponential-like distribution $E(w_{\text{min}}, w_{\text{max}})$ \cite{yang2021cnn}.
Finally, design rules are guaranteed via rejection sampling (lines~13-14). 

\begin{algorithm}[t]
\caption{Synthetic Training Data Generation.}\label{algo:synthetic}
\KwIn{Process technology $\mathcal{T}$, expected conductor count $N$, window width $W$.}
\KwOut{A list of randomly placed conductors $\bm{\mathcal{S}}$.}
Randomly select a \texttt{LayerSet} of $U\{3,4,5\}$ layers in $\mathcal{T}$\;
Sample $n\sim P(N-2) + 2$; \hfill \textcolor{gray}{$\rhd$ensure $n\ge2$ and $\mathbb{E}(n)=N$}\\
$\bm{\mathcal{S}}\gets [\texttt{substrate}]$; \hfill \textcolor{gray}{$\rhd$init with an infinite substrate}\\
\Repeat{$\bm{\mathcal{S}}$ has $n$ conductors}{
    $(y,h,w_{\text{min}},s_{\text{min}})\gets$ sample a layer from \texttt{LayerSet}\;
    Sample $x\sim \mathcal{N}(0,(\frac{W}{6})^2)$; \hfill \textcolor{gray}{$\rhd 3\sigma\text{ range}=W$}\\
    $w_{\text{max}}\gets W-2|x|$\;
    \If{$U(0,1) < 0.1$}{
        Sample $w \sim U(w_{\text{min}}, w_{\text{max}})$\;
    }\Else{
        Sample $w \sim E(w_{\text{min}}, w_{\text{max}})$\;
        $x\!\gets$round $x$ to grids $i(w_{\text{min}}\!+\!s_{\text{min}}), i\!=\!0,\pm 1\dots$\;
    }
    \If{$\texttt{DesignRuleCheck}(x,y,w,h, \bm{\mathcal{S}})$ passes}{
        Add $(x,y,w,h)$ to $\bm{\mathcal{S}}$\;
    }
}
\end{algorithm}

In our implementation, $N$ is set to 8, resulting in roughly 2-16 conductors per sample. The window width $W$ is technology dependent and determined via field-solver simulation such that the coupling between two M1 conductors with spacing $W/2$ falls below $1\%$ of the total capacitance, following common practice \cite{yang2021cnn,abouelyazid2022fast}. For instance, $W$ is 10.7$\mu$m for a 65nm node and 2.7$\mu$m for a 7nm node.
We implement $E(w_{\text{min}}, w_{\text{max}})$ as $\mathbb{P}(w\!=\!iw_{\text{min}})\propto (0.75)^{i},i\!=\!1,2,\dots$, truncated to $w_{\text{max}}$, which is similar to \cite{yang2021cnn}.

\section{Experimental Results}
\subsection{Experiment Setup}

All experiments were conducted on Intel Xeon Platinum 8488C CPUs (3.8GHz) and a single NVIDIA RTX 5090 GPU.
Synthetic data generation (Alg.~\ref{algo:synthetic}) and 2D field solver were implemented in C++ with CPU multithreading.
Model training and evaluation used PyTorch 2.9 and CUDA 13.0.

\begin{figure}[t]
    \centering
\setlength{\abovecaptionskip}{2pt}
\setlength{\belowcaptionskip}{2pt}
    \includegraphics[width=\linewidth]{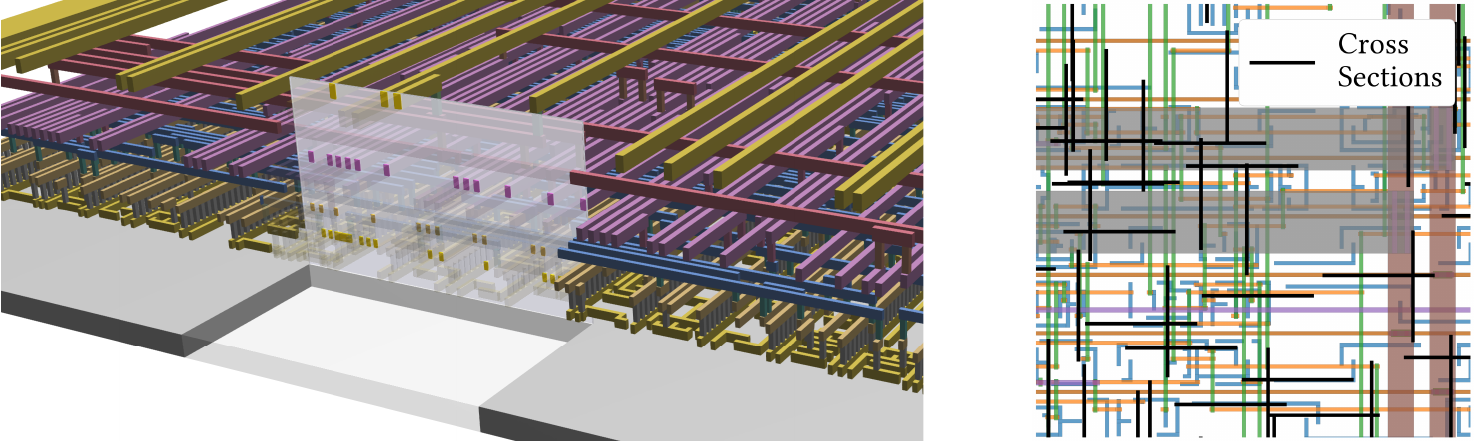}
    \caption{Example cross-sections from ASAP7 real-design test set, in local 3D view (left) and local top view (right).}
    \label{fig:cross_section}
\end{figure}

\textbf{Datasets}. We use two open-source process nodes (ASAP7 \cite{clark2016asap7} and FreePDK15 \cite{bhanushali2015freepdk15}) and two industrial nodes that include conformal dielectrics (Real28 and Real65). For each node, we generate 50K synthetic cross-sections using Alg.~\ref{algo:synthetic} and solve capacitance matrices with 2D field solver, 
which takes about 16 hours on 24 threads. 
The synthetic datasets are split 9:1 into training/validation sets. For testing, 
we prepare three designs for ASAP7 and Real65, respectively, covering multiple circuit types. We uniformly sample 5K cross-sections from each design and retain the 10 center-most conductors in each sample, as shown in Fig.~\ref{fig:cross_section}. FreePDK15 and Real28 include only synthetic data for training the multi-node model.

\textbf{Models}. 
We implement \textbf{AttentionCap} with $d=256, d_k=64, M=4, d_{\text{ff}}=512, L=6$ (4.0M parameters).
For large multi-node datasets, we also evaluate \textbf{AttentionCap-L} with $d=384, d_k=96, M=4, d_{\text{ff}}=768, L=8$ (12M parameters).
As a baseline, \textbf{AttentionCap w/o attention} (equivalent to a pure \textbf{MLP}) removes all attention modules and expands to $d=512, d_{\text{ff}}=1024$ (9.7M parameters). We include \textbf{CNN-Cap} \cite{yang2021cnn} from its official implementation \cite{cnncap}, which is a ResNet34-based model with 7.2M parameters.

\textbf{Training}. 
All models are trained under a unified setup: AdamW optimizer with $\beta_1=0.9, \beta_2=0.999$ and weight decay $10^{-4}$; learning rate scheduled from $1.5\!\times\!10^{-4}$ to $10^{-5}$ with 1K warm-up steps and linear decay.
We apply horizontal-flip augmentation for training data (simply $x_i\mapsto -x_i$) and the proposed normalized Laplacian loss.
We consistently train for 300K steps and select the best validation-loss checkpoint. To maintain roughly 400 effective epochs for different datasets, the training batch size is adjusted according to:
$\text{batch\_size}\times \text{steps}=\text{data\_size}\times\text{epochs}$. For instance, batch size is 128 for training on the 50K dataset, 512 for 200K. During testing, we apply a batch size of 128 across all experiments.

\textbf{Evaluation Metrics}. Following \cite{yang2021cnn}, we report four key metrics in practical capacitance extraction:
Err$_\text{tot}$, Ratio$_\text{tot}$, Err$_\text{cp}$, and Ratio$_\text{cp}$. 
Denote $e^{(k)}_{ij}=\left|\hat{C}^{(k)}_{ij}-C^{(k)}_{ij}\right| / \left|{C^{(k)}_{ij}}\right|$ for the $k$-th sample. 
Err$_{\text{tot}}$ and Ratio$_\text{tot}$ are the mean and high-error rate ($>$5\%) of total-capacitance errors $e^{(k)}_{ii},\forall k,i$;  
Err$_{\text{cp}}$ and Ratio$_\text{cp}$ are the mean and high-error rate ($>$10\%) of coupling-capacitance errors $e^{(k)}_{ij},\forall k,i\ne j$. Tiny couplings (less than 1\% of its total capacitance) are excluded when calculating these metrics.

\subsection{Benchmark on CNN-Cap Public Datasets}
We first benchmark AttentionCap on CNN-Cap datasets \cite{yang2021cnn} available at \cite{cnncap}, which contain fixed 3-layer combinations at 55nm and 15nm nodes, with 50K cross-sections per setting.
Given the density-grid data,
we reconstruct each conductor's $\tilde{x}_i, \tilde{w}_i$ based on grid index and use $(\tilde{x}_i, \tilde{w}_i,\text{layer\_id})$ as AttentionCap inputs. The datasets provide single-master capacitances per cross-section, so we train and test AttentionCap on a single row of its output matrix. Validation-set errors are shown in Table~\ref{tab:cnncap}. AttentionCap consistently outperforms CNN-Cap with remarkable accuracy.
\begin{table}[h]
    \centering
\setlength{\abovecaptionskip}{2pt}
\setlength{\belowcaptionskip}{2pt}
    \caption{Benchmark on CNN-Cap Pattern-C datasets \cite{yang2021cnn,cnncap}.}
    \label{tab:cnncap}
    \setlength\tabcolsep{3.5pt}
    \begin{threeparttable}
\begin{tabular}{l c c cccc}
    \toprule
    Model & Node & \begin{tabular}{@{}c@{}}Layers\end{tabular} & Err$_\text{tot}$ & Ratio$_\text{tot}$ &
    Err$_\text{cp}$  & Ratio$_\text{cp}$ \\
    \midrule
    CNN-Cap\tnote{$\ast$} & \multirow{2}{*}{55\,nm} & \multirow{2}{*}{(2,3,6)} & 0.10\% & \textbf{0} & 1.20\% & 0.30\% \\
    AttentionCap & & & \textbf{0.03\%} & \textbf{0} & \textbf{0.35\%} & \textbf{0.02\%} \\
    \midrule
    CNN-Cap\tnote{$\ast$} & \multirow{2}{*}{55\,nm} & \multirow{2}{*}{(2,4,6)} & 0.20\% & \textbf{0} & 1.20\% & 0.10\% \\
    AttentionCap & & & \textbf{0.04\%} & \textbf{0} & \textbf{0.32\%} & \textbf{0.03\%} \\
    \midrule
    CNN-Cap\tnote{$\ast$} & \multirow{2}{*}{15\,nm} & \multirow{2}{*}{(1,3,5)} & 0.20\% & \textbf{0} & 1.80\% & 0.40\% \\
    AttentionCap & & & \textbf{0.03\%} & \textbf{0} & \textbf{0.29\%} & \textbf{0} \\
    \midrule
    CNN-Cap\tnote{$\ast$} & \multirow{2}{*}{15\,nm} & \multirow{2}{*}{(1,3,8)} & 0.20\% & \textbf{0} & 1.50\% & \textbf{0} \\
    AttentionCap & & & \textbf{0.02\%} & \textbf{0} & \textbf{0.30\%} & \textbf{0} \\
    \bottomrule
\end{tabular}
  \begin{tablenotes}[flushleft]
  \small
    \item[$\ast$] Results are taken from \cite{yang2021cnn}. CNN-Cap uses separate models for total and coupling capacitance; they are merged into one row here.
  \end{tablenotes}
\end{threeparttable}
\end{table}

\subsection{Main Results: Training on Synthetic Data and Testing on Real Designs}
Our principle is to train models on synthetic data and test them on unseen real-design cross-sections, under \textbf{multi-metal-layer} and \textbf{multi-process-node} settings.

\begin{table*}[t]
    \centering
      \renewcommand{\arraystretch}{0.9}
	\setlength{\abovecaptionskip}{1pt}
    \setlength{\belowcaptionskip}{1pt}
    \caption{Main Results: training on synthetic data and testing on unseen real designs.}
    \label{tab:main}
    \setlength\tabcolsep{2.5pt}
\begin{threeparttable}
    \begin{tabular}{lc c cccc cccc ccccc}
    \toprule
    \tworows{Model} & \tworows{Param.} &
    \tworows{Node} &
    \multicolumn{4}{c}{Valid. Error (synthetic data)} &
    \multicolumn{4}{c}{Test Error (real-design)} &
    \tworows{\begin{tabular}{@{}c@{}}\#Train\\Samples\tnote{a}\end{tabular}} &
    \tworows{\begin{tabular}{@{}c@{}}Train\\Time\tnote{b}\end{tabular}} &
    \tworows{\begin{tabular}{@{}c@{}}\#Test\\Samples\tnote{a}\end{tabular}} &
    \tworows{\begin{tabular}{@{}c@{}}Test\\Time\tnote{b}\end{tabular}} &
    \tworows{\begin{tabular}{@{}c@{}}Test\\FLOPs\end{tabular}} \\
    \cmidrule(lr){4-7} \cmidrule(lr){8-11}
    & & & 
    Err$_\text{tot}$ & Ratio$_\text{tot}$ &
    Err$_\text{cp}$  & Ratio$_\text{cp}$  &
    Err$_\text{tot}$ & Ratio$_\text{tot}$ &
    Err$_\text{cp}$  & Ratio$_\text{cp}$  &
    & & & \\
    \midrule
CNN-Cap$_\text{tot}$ & \multicolumn{1}{c}{7.2M} & \multirow{4}{*}{65\,nm}
    & \multicolumn{1}{|c}{2.83\%} & 15.0\% & - & \multicolumn{1}{c|}{-}
    & 4.06\% & 30.0\% & - & \multicolumn{1}{c|}{-}
    & 315\,K  & 2.18\,hr & 54\,K & 6.91\,s & 9.83\,B \\
CNN-Cap$_\text{cp}$ & \multicolumn{1}{c}{7.2M} & 
    & \multicolumn{1}{|c}{-} & - & 9.74\% & \multicolumn{1}{c|}{34.3\%}
    & - & - & 20.4\% & \multicolumn{1}{c|}{54.4\%}
    & 1265\,K & 9.69\,hr & 184\,K & 27.6\,s & 33.1\,B \\
MLP & \multicolumn{1}{c}{9.7M} & 
  & \multicolumn{1}{|c}{14.5\%} & 76.2\% & 53.7\% & \multicolumn{1}{c|}{81.9\%}
  & 23.6\% & 87.8\% & 84.0\% & \multicolumn{1}{c|}{87.0\%}
  & 45\,K & 0.43\,hr  & 15\,K & 0.15\,s & 0.67\,B \\
  AttentionCap & \multicolumn{1}{c}{4.0M} & 
  & \multicolumn{1}{|c}{\textbf{0.85\%}} & \textbf{1.65\%} & \textbf{3.24\%} & \multicolumn{1}{c|}{\textbf{6.95\%}}
  & \textbf{0.98\%} & \textbf{1.55\%} & \textbf{5.38\%} & \multicolumn{1}{c|}{\textbf{13.4\%}}
  & 45\,K & 0.64\,hr & 15\,K& 0.28\,s & 0.28\,B \\
\midrule

CNN-Cap$_\text{tot}$ & \multicolumn{1}{c}{7.2M} & \multirow{4}{*}{7\,nm}
  & \multicolumn{1}{|c}{0.69\%} & \textbf{0.09\%} & - & \multicolumn{1}{c|}{-}
  & 2.11\% & 9.94\% & - & \multicolumn{1}{c|}{-}
  & 315\,K & 2.18\,hr & 73\,K & 11.1\,s & 13.2\,B \\

CNN-Cap$_\text{cp}$ & \multicolumn{1}{c}{7.2M} & 
  & \multicolumn{1}{|c}{-} & - & 3.89\% & \multicolumn{1}{c|}{6.95\%}
  & - & - & 25.1\% & \multicolumn{1}{c|}{48.7\%}
  & 1416\,K & 9.61\,hr & 322\,K & 54.0\,s & 58.2\,B \\

MLP & \multicolumn{1}{c}{9.7M} & 
  & \multicolumn{1}{|c}{16.4\%} & 80.9\% & 62.5\% & \multicolumn{1}{c|}{85.1\%}
  & 32.4\% & 90.8\% & 120\% & \multicolumn{1}{c|}{86.8\%}
  & 45\,K & 0.44\,hr & 15\,K & 0.15\,s & 0.86\,B \\

AttentionCap & \multicolumn{1}{c}{4.0M} & 
  & \multicolumn{1}{|c}{\textbf{0.47\%}} & 0.39\% & \textbf{2.19\%} & \multicolumn{1}{c|}{\textbf{3.49\%}}
  & \textbf{1.52\%} & \textbf{6.75\%} & \textbf{7.13\%} & \multicolumn{1}{c|}{\textbf{18.9\%}}
  & 45\,K & 0.64\,hr & 15\,K & 0.24\,s & 0.35\,B \\

\midrule

MLP & \multicolumn{1}{c}{9.7M} & \multirow{3}{*}{\textbf{Mix}}
  & \multicolumn{1}{|c}{15.2\%} & 78.4\% & 57.9\% & \multicolumn{1}{c|}{83.7\%}
  & 30.0\% & 90.0\% & 110\% & \multicolumn{1}{c|}{86.7\%}
  & 180\,K & 0.52\,hr & 30\,K & 0.10\,s & 1.53\,B \\
AttentionCap & \multicolumn{1}{c}{4.0M} & 
  & \multicolumn{1}{|c}{0.54\%} & 0.46\% & 2.55\% & \multicolumn{1}{c|}{4.61\%}
  & 0.88\% & 1.07\% & 5.25\% & \multicolumn{1}{c|}{13.5\%}
  & 180\,K & 0.68\,hr & 30\,K & 0.47\,s & 0.63\,B \\
AttentionCap-L & \multicolumn{1}{c}{12M} & 
  & \multicolumn{1}{|c}{\textbf{0.44\%}} & \textbf{0.30\%} & \textbf{2.08\%} & \multicolumn{1}{c|}{\textbf{2.97\%}}
  & \textbf{0.67\%} & \textbf{0.28\%} & \textbf{3.99\%} & \multicolumn{1}{c|}{\textbf{9.75\%}}
  & 180\,K & 1.02\,hr & 30\,K & 0.66\,s & 1.89\,B \\
  
    \bottomrule
\end{tabular}
\begin{tablenotes}
    \small
    \item[a] All models use the same train/valid/test data. The different sample sizes are due to architectural difference: AttentionCap treats each matrix as a sample, while CNN-Cap treats each capacitance element as an individual sample. We increase CNN-Cap batch size accordingly to align training steps/epochs.
    \item[b] CNN-Cap runtimes also include the cost of constructing density-grid inputs, which is negligible for training but accounts for $\sim$74\% of testing time.
\end{tablenotes}
\end{threeparttable}

\end{table*}

\textbf{Single-Node Learning}. For ASAP7 and Real65, we have 50K synthetic samples and 15K test samples each. To train CNN-Cap \cite{yang2021cnn}, we discretize each metal layer into a 512-resolution density grid and expand its input channels to 9 (ASAP7) and 8 (Real65) to handle all metal layers. CNN-Cap$_{\text{tot}}$ and CNN-Cap$_{\text{cp}}$ are the specialized models for total and coupling capacitance \cite{yang2021cnn}. 
Main results are summarized in Table~\ref{tab:main}. AttentionCap significantly outperforms CNN-Cap and MLP baselines on both nodes, achieving 0.98\%/5.38\% total/coupling error on 65nm and 1.52\%/7.13\% total/coupling error on 7nm,
in unseen real-design tests. CNN-Cap is relatively accurate on total capacitance but suffers from large coupling error ($>$20\% average error; $\sim$50\% high-error outliers), showing its limited ability to capture complex multi-conductor interactions. MLP (i.e., AttentionCap w/o attention) fails to provide usable predictions, highlighting the fundamental role of attention mechanism.
Test errors are consistently higher than validation errors, indicating the challenging \textbf{out-of-distribution} (OOD) features in real designs.
Moreover, models show higher error than on 3-layer datasets (Table~\ref{tab:cnncap}), which confirms that our multi-layer setting presents much higher learning complexity.

Moreover, AttentionCap is far more efficient than CNN-Cap: 18.5$\times$ faster in training and 192$\times$ faster (181$\times$ fewer FLOPs) in testing, owing to smaller model size and avoidance of density-grid processing. 
For comparison, field solver takes 6 hours on 24 threads to solve all test samples.

\begin{figure}[b!]
    \centering
    \setlength{\abovecaptionskip}{1pt}
    \setlength{\belowcaptionskip}{1pt}
    \includegraphics[width=\linewidth]{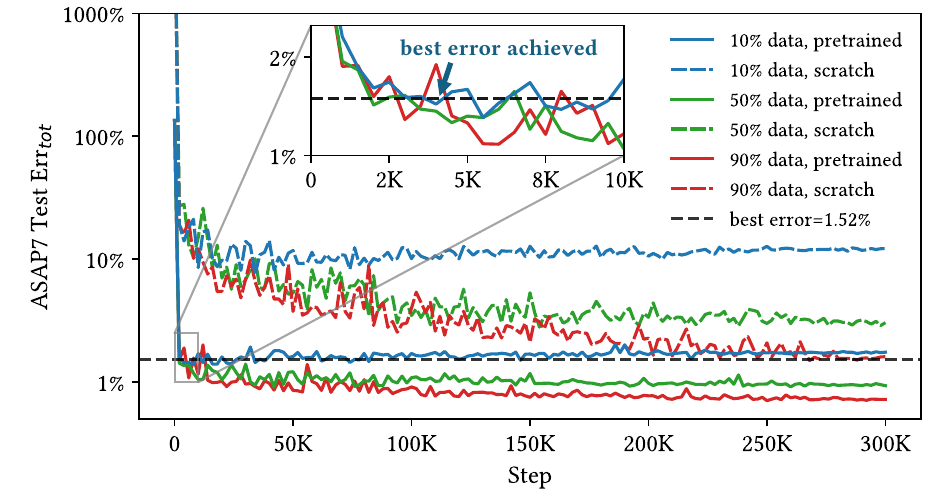}
    \caption{Strong transferability of AttentionCap: with only 5K samples and 4K finetuning steps for new process node (7nm), the pretrained model (on 15nm+28nm+65nm) outperforms the best model trained from scratch.}
    \label{fig:transfer}
\end{figure}
\textbf{Multi-Node Learning}. We mix 7nm, 15nm, 28nm, 65nm synthetic samples to train multi-node models, and test them on 7nm and 65nm unseen real designs. 
As shown in Table~\ref{tab:main}, AttentionCap achieves even higher 7nm/65nm test accuracy than models trained only on 7nm/65nm, which implies that AttentionCap can accurately distinguish samples via process-node embeddings, and has been enhanced by samples from other nodes via learning node-agnostic features.
The larger AttentionCap-L further improves accuracy given the abundant data, attaining the best \textbf{0.67\%/3.99\% test error} for total/coupling capacitance, which is 4.6$\times$/5.7$\times$ lower than CNN-Cap's average error.

The learned node-agnostic features make AttentionCap highly transferable to new nodes.
We pretrain a multi-node AttentionCap on 15nm+28nm+65nm data, and finetune it with 10\%/50\%/90\% train splits from 7nm data.
As shown in Fig.~\ref{fig:transfer}, with only a 10\% train split (5K samples) and 4K finetuning steps, the pretrained model outperforms the best 7nm model trained from scratch (best error achieved at 240K steps with the 90\% train split). 
This data-efficient adaption is exceptionally valuable in practice, as preparing training data 
is the main bottleneck.
We attribute this multi-node capability to the exceptional representation power of Transformer architecture.

\subsection{Ablation Study}
We conduct an ablation study with the best model AttentionCap-L on the mixture dataset (7nm+15nm+28nm+65nm). 
We evaluate training with mean squared error (MSE) and a GELU-based feed-forward network (FFN) variant (using $d_\text{ff}=1152$ to match 12M parameters). Test-set errors are reported in Table~\ref{tab:ablation}, showing that the proposed normalized Laplacian loss is crucial for training, and the SwiGLU \cite{shazeer2020glu} improves capacitance accuracy.
\begin{table}[h]
    \centering
\setlength{\abovecaptionskip}{1pt}
\setlength{\belowcaptionskip}{1pt}
    \caption{Ablation study on AttentionCap-L (12M).}
\label{tab:ablation}
    \setlength\tabcolsep{4pt}
\begin{tabular}{c l l cccc}
    \toprule
    Node & FFN & Loss & Err$_\text{tot}$ & Ratio$_\text{tot}$ &
    Err$_\text{cp}$  & Ratio$_\text{cp}$ \\
    \midrule
    \multirow{3}{*}{\textbf{Mix}} 
    & SwiGLU &MSE & 4.91\% & 34.8\% & 25.6\% & 49.7\% \\
    & GELU & Laplacian & 1.00\% & 0.89\% & 5.75\% & 15.1\% \\
     & SwiGLU & Laplacian & \textbf{0.67\%} & \textbf{0.28\%} & \textbf{3.99\%} & \textbf{9.75\%} \\

    \bottomrule
\end{tabular}
\end{table}

\section{Conclusion}
We introduce AttentionCap, a Transformer for capacitance matrix learning with Gram representation learning, symmetric-attention output, and normalized Laplacian loss. AttentionCap delivers high accuracy, broad generalization across metal layers and process nodes.
Trained exclusively on synthetic data, it achieves 0.67\%/3.99\% self/coupling error on unseen real designs, with 4.6$\times$/5.7$\times$ lower error and 192$\times$ higher efficiency than CNN-Cap \cite{yang2021cnn}. It also adapts to a new node with only 5K samples and a quick finetune.
These results demonstrate AttentionCap’s strong potential as a unified and transferable model for practical capacitance extraction.
\newpage
\bibliographystyle{ACM-Reference-Format}
\bibliography{sample-base}

\end{document}